\documentclass[10pt,twocolumn,letterpaper]{article}

\usepackage[pagenumbers]{iccv} 

\definecolor{iccvblue}{rgb}{0.21,0.49,0.74}
\usepackage[pagebackref,breaklinks,colorlinks,allcolors=iccvblue]{hyperref}
\usepackage{pifont}

\def\paperID{10880} 
\def\confName{ICCV}
\def\confYear{2025}

\title{MagShield: Towards Better Robustness in Sparse Inertial Motion Capture Under Magnetic Disturbances}

\author{
    Yunzhe Shao\textsuperscript{1}, 
    Xinyu Yi\textsuperscript{1}, 
    Lu Yin\textsuperscript{2}, 
    Shihui Guo\textsuperscript{2}, 
    Junhai Yong\textsuperscript{1},
    Feng Xu\textsuperscript{1}\\[0.2cm]
    \normalsize{\textsuperscript{1}School of Software and BNRist, Tsinghua University} \\ 
    \normalsize{\textsuperscript{2}School of Informatics, Xiamen University}
}

\begin{document}

\twocolumn[{%
\renewcommand\twocolumn[1][]{#1}%
\maketitle
\includegraphics[width=\linewidth]{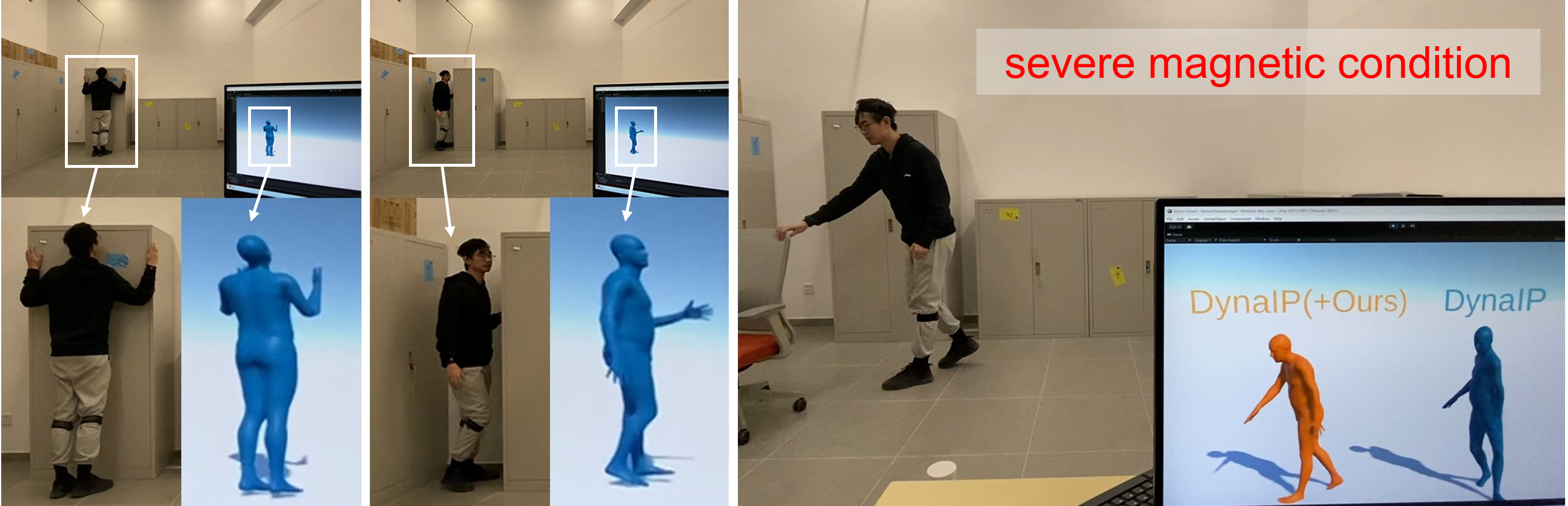}
\captionof{figure}{We propose MagShield, an IMU orientation estimation method that specifically designed to address magnetic interference in sparse inertial MoCap systems. By integrating with our method, existing sparse inertial MoCap systems (left, PNP+ours; right, DynaIP+ours) can operate reliably even under severe magnetic condition. \vspace{1em}}
\label{fig:teaser}
}]

\begin{abstract}
This paper proposes a novel method called MagShield, designed to address the issue of magnetic interference in sparse inertial motion capture (MoCap) systems. Existing Inertial Measurement Unit (IMU) systems are prone to orientation estimation errors in magnetically disturbed environments, limiting their practical application in real-world scenarios. To address this problem, MagShield employs a ``detect-then-correct'' strategy, first detecting magnetic disturbances through multi-IMU joint analysis, and then correcting orientation errors using human motion priors. MagShield can be integrated with most existing sparse inertial MoCap systems, improving their performance in magnetically disturbed environments. Experimental results demonstrate that MagShield significantly enhances the accuracy of motion capture under magnetic interference and exhibits good compatibility across different sparse inertial MoCap systems. Code and dataset will be released.
\end{abstract}    
\section{Introduction}
\label{sec:intro}

Motion capture is widely used in animation, virtual reality, and embodied-AI. 
Among various motion capture systems, IMUs (Inertial Measurement Units)-based solutions have gained increasing popularity. 
Early inertial MoCap systems \cite{noitom, xsens} adopt a dense configuration, requiring actors to wear 17 IMUs. 
While these approaches achieve promising results and have been commercialized, the excessive number of IMUs causes inconvenience for users. 
To achieve low cost and portability, some researchers have explored the use of sparser IMU configurations with 6 IMUs. 
The continuous advancement has enabled these methods \cite{DIP, TransPose, PIP, TIP, PNP} to achieve notable progress in real-time performance, accuracy, and physical plausibility.

However, a critical challenge has been neglected in previous work, i.e., IMUs are susceptible to orientation errors in magnetically disturbed environments \cite{fan2017magnetic}. 
This limitation severely restricts the practical use of present IMU-based MoCap systems, confining them to environments with stable magnetic fields and hindering their applicability in real-world scenarios such as indoor spaces or near electronic devices. 
The root cause of this issue is that IMUs rely on magnetometer measurements, which are assumed to represent the Earth's magnetic field, to estimate their global orientation. 
Sometimes, when magnetic disturbances are present, the magnetometer captures these anomalies without recognizing them as interference, mistakenly interpreting the noisy magnetic field as the true geomagnetic field. 
This leads to incorrect orientation estimates, further degrading the performance of inertial MoCap systems.

In this work, we propose MagShield, the first method designed to alleviate the effects of \textbf{\emph{mag}}netic interference in sparse IMU-based MoCap systems, acting as a protective \textbf{\emph{shield}}. Our approach can be easily integrated with various sparse inertial MoCap systems to enhance their performance in magnetically disturbed environments. 
Specifically, we address the problem by a detect-then-correct strategy: (1) improving the accuracy of detecting ``disturbed magnetic fields'', and (2) correcting orientation errors when they occur, providing a post-hoc remedy. 
The fundamental idea of our method is to leverage the prior knowledge that ``IMUs are worn on the human body". 
In the following, we detail how this prior is applied in each stage to tackle magnetic interference.

In the disturbance detection stage, our method evaluates the ambient magnetic field by jointly analyzing measurements of multiple adjacent IMUs. 
Unlike previous approaches \cite{zhang2012quaternion,sabatini2011estimating,madgwick2020extended} which rely on single IMU readings, our method assesses the spatial stability of the magnetic field in a local region with multiple IMUs. 
Thanks to the body-worn nature of IMUs, we can easily compute the relative positional relationships between IMUs based on the human pose, which are used to intergrating the measurements of multiple IMUs. 
In our implementation, we employ an auto-regressive framework that incorporates previously estimated poses into the magnetic field assessment process. 
This enables more accurate detection of distortions, thereby reducing IMU errors and improving the precision of motion capture.

In the error correction stage, we correct orientation errors also by utilizing human motion priors. 
Specifically, when IMUs exhibit errors, their readings often correspond to implausible motions, allowing us to detect and correct such errors. 
While previous works \cite{TransPose, PNP} trained networks on noisy data to develop general noise resistance, we design a dedicated network specifically for magnetic interference by leveraging its unique noise model. 
Training such a denoising network requires a large dataset of IMU data affected by magnetic distortions, which is difficult to obtain. 
To address this, we propose a novel data synthesis method that simulates the impact of magnetic fields on IMU readings.

In summary, our main contributions include:
\begin{itemize}
\item MagShield, the first method specifically designed to address magnetic interference in sparse inertial MoCap systems, which can be integrated with most existing MoCap methods to enhance their robustness against magnetic disturbances.
\item A pose-aware magnetic disturbance detection method that improves the accuracy of detecting corrupted magnetic fields with multiple body-worn IMUs.
\item A motion prior-based error correction network, leveraged by a novel data synthesis method, to correct orientation errors caused by magnetic interference.
\end{itemize}

\begin{figure*}
    \centering
    \includegraphics[width=1\linewidth]{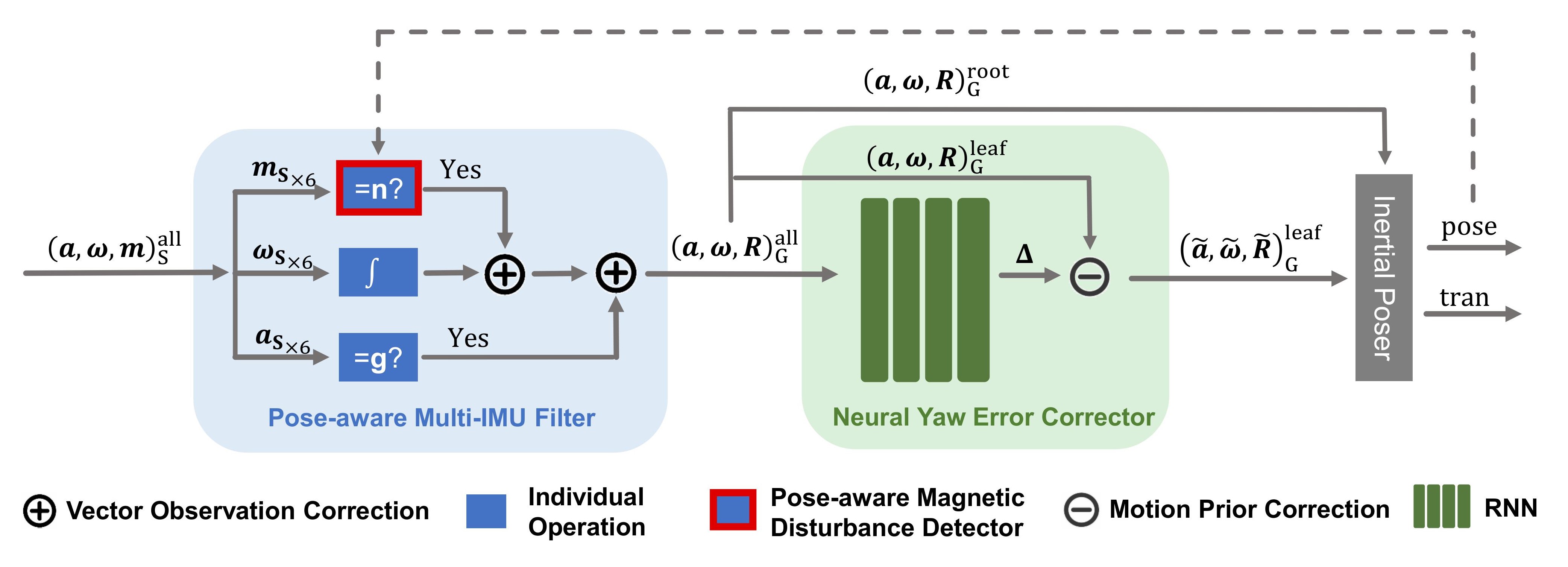}
    \caption{Overview of our method. In the first stage (blue, left), we transform sensor-local measurements (subscript S) to global readings (subscript G). A pose-aware magnetic disturbance detector is used for reject disturbed magnetic field. In the second stage (green, right), we further refine the leaf IMU readings using human motion prior. The output IMU readings of our method are then fed into inertial posers.}
    \label{fig:pipeline}
\end{figure*}

\section{Related Works}
\subsection{Motion Capture from Inertial Sensors}
An IMU (Inertial Measurement Unit) is a lightweight sensor capable of measuring orientation and acceleration. When attached to the human body, it can track bone rotations and reconstruct human motion. Commercial inertial motion capture systems\cite{noitom, xsens} typically employ a dense configuration, attaching 17 IMUs to the body to fully determine the rotation of each bone. While accurate, the excessive number of sensors makes the system cumbersome to wear and limits mobility. To this end, sparse IMU motion capture has remained an active area of research. The pioneering work SIP \cite{SIP} demonstrates the feasibility of using six IMUs for pose estimation, but it relies on offline optimization and cannot achieve real-time performance. DIP \cite{DIP} first achieves real-time human pose estimation, while TransPose \cite{TransPose} further advances it by realizing real-time estimation of both pose and translation. Subsequent approaches have explored various perspectives to enhance accuracy. PIP \cite{PIP} improved the physical plausibility of the results by incorporating physical constraints. TIP \cite{TIP} addressed the interaction between humans and terrain, enabling motion capture in non-flat environments. PNP \cite{PNP} and DynaIP \cite{DynaIP} tackle the utilization of acceleration data from different perspectives: while PNP introduce non-inertial forces to resolve ambiguities in acceleration under non-inertial frames, DynaIP found that leveraging acceleration to estimate velocity first is more beneficial for network-based learning of motion characteristics. ASIP \cite{ASIP} employed a Sequence Structure Module for better capturing spatial correlations between joints. Additionally, some studies have begun to explore the use of fewer IMUs \cite{IMUPoser,DiffPoser,MobilePoser}. Despite the rapid development of IMU-based motion capture in various aspects, a critical issue has long been sidestepped, that is the interference of magnetic fields on IMUs. In this work, we propose a method to enhance the resilience of sparse IMU motion capture systems to magnetic disturbances.

\subsection{Magnetic Field Disturbance Detection}
Magnetic field disturbance detection plays a crucial role in IMU orientation estimation. The existing approaches can be categorized into threshold-based methods and model-based methods. Threshold-based methods rely on the observation that magnetic interference often leads to noticeable deviations in magnetic field characteristics. Thresholds are typically set for the magnetic field strength \cite{zhang2012quaternion,sabatini2011estimating,madgwick2020extended}, the dip angle \cite{yadav2014accurate}, or both \cite{lee2009minimum, fan2017adaptive, laidig2023vqf} to detect interference. This solution is easy to implement; however, the threshold tuning process is usually troublesome. Another category of methods is model-based approaches, which regard magnetic field disturbances as state variables that can be estimated through probabilistic model. The earliest model-based work \cite{roetenberg2005compensation} approximated magnetic field as a first-order Markov process. However, this approach assumes a fixed disturbance pattern and fails to ensure reliable performance in mismatched scenarios. An adaptive approach is proposed by \cite{sabatini2012variable}, which switches between different magnetic field models according to the noise level, thereby achieving more robust noise estimation. Although model-based methods eliminate the need for parameter tuning, their complex implementation and significant computational overhead have limited their widespread adoption. It should be noted that, both of the threshold-based method and the model-based method rely solely on information from a single IMU, overlooking the potential of leveraging multiple IMUs in motion capture scenarios. When IMUs are worn on the human body, their relative positions and orientations can provide additional information for detecting magnetic disturbances, offering a more robust solution.
\section{Preliminaries}

Our task is to mitigate the impact of magnetic interference on IMU-based MoCap systems. To lay the groundwork, we first explain how IMUs work, then introduce how magnetic fields affect IMU orientation estimation.

An IMU consists of an accelerometer, a gyroscope and a magnetometer. Their sensor-local measurements, including acceleration $\boldsymbol a$, angular velocity $\boldsymbol \omega$, and magnetic field vector $\boldsymbol m$, can be used to estimate IMU's global orientation $\boldsymbol R$. This process relies on two core algorithms: angular velocity integration and vector observation (VO) correction. Angular velocity can be viewed as the change in sensor orientation between two consecutive frames. Ideally, the sensor's orientation can be calculated by integrating the angular velocity. However, due to sensor noise \cite{bhardwaj2018errors} and numerical integration errors, the orientation error obtained through this method accumulates over time. To address this limitation, VO correction is often introduced. The core idea of the VO correction is that by observing certain global vectors in the sensor-local coordinate frame, the current orientation of the sensor can be inferred. Accelerometer and magnetometer are often used for VO correction, as they can sense gravitational acceleration and the geomagnetic field, respectively. The combination of the two VO corrections and angular velocity integration through filtering algorithms typically yields accurate results.

However, in environments with magnetic interference, using magnetometer for VO correction may have counterproductive effects. This is because it can be misled by the disturbed magnetic fields. In contrast, accelerometer measurements are generally more reliable. Therefore, researchers have approached the problem of magnetic interference from two perspectives. On the one hand, various methods have been developed to detect whether the magnetic field is being interfered with, and this is precisely the problem that the first stage of our method aims to solve. On the other hand, they propose decoupling the estimation of attitude (roll and pitch) and heading (yaw), where gravitational VO is responsible for the former, and magnetic VO handles the latter. This can be done through different ways \cite{zhang2016dual, madgwick2020extended, suh2012quaternion, shuster1981three}. As a result, magnetic disturbances primarily affect yaw, enabling orientation errors to be described as a single degree of freedom, termed ``yaw error'' \cite{bachmann2004investigation}. The design of the second stage of our method is based on this characteristic.

\section{Method}

Our method is an IMU orientation estimation module designed to enhance the resilience of sparse inertial MoCap systems to magnetic interference. The inputs to our system are the \emph{sensor-local} raw measurements (including  accelerations, angular velocities, and magnetic field) of 6 IMUs placed on forearms, lower legs, head and the root. It outputs processed IMU readings including \emph{global} accelerations, angular velocities, and orientations, which are then fed into inertial posers for human motion estimation. To avoid ambiguity, throughout the entire article, ``raw measurements'' refer to the \emph{sensor-local} information, which can be directly measured. In contrast, ``readings'' refer to \emph{global} information, which are derived from raw measurements. As shown in Fig. \ref{fig:pipeline}, our approach comprises two stages: (1) a pose-aware multi-IMU fusion algorithm, which aims to derive IMU readings from their raw measurements (Sec. \ref{sec:Pose-aware Multiple IMUs Fusion}), and (2) a neural yaw error corrector, which further refines the readings of leaf IMUs (Sec. \ref{sec:Neural Yaw Error Corrector}). 

\subsection{Pose-aware Multi-IMU Filter}
\label{sec:Pose-aware Multiple IMUs Fusion}
\begin{figure}
    \centering
    \includegraphics[width=1\linewidth]{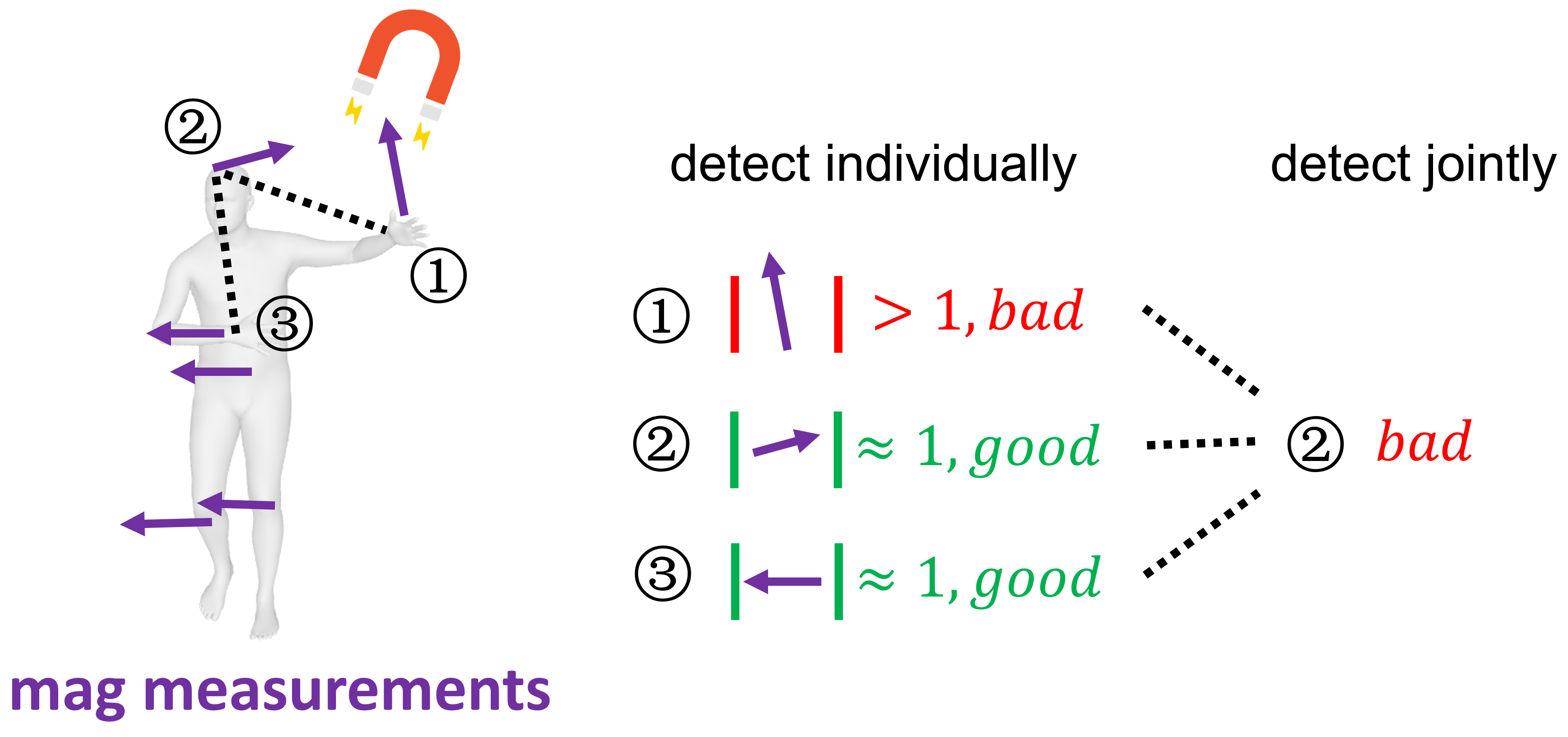}
    \caption{Illustration to our magnetic disturbance detector. Now the magnetic fields at both \ding{192} and \ding{193} are disturbed. However, the magnitude at \ding{193} happens to be close to 1. If each IMU is considered individually, \ding{193} won't be detected. However, when \ding{193} aggregate its neighbors' information, the disturbance at \ding{193} will be detected.}
    \label{fig:detector}
\end{figure}

In this stage, we perform fusion on the raw IMU measurements. The inputs of this stage are raw measurements of 6 IMUs, denoted as $(\boldsymbol a, \boldsymbol \omega, \boldsymbol m)_\mathrm{S}^\mathrm{all}$, the outputs are IMU readings $(\boldsymbol a, \boldsymbol \omega, \boldsymbol R)_\mathrm{G}^\mathrm{all}$. The core challenge lies in accurately determining whether the current magnetic field should be used for VO correction. We first describe how we utilize magnetometer data from multiple IMUs to detect and reject disturbed magnetic fields (Sec. \ref{sec:Pose-aware Magnetic Field Selector}), then explain how this information is fused with accelerometer and gyroscope measurements (Sec. \ref{sec:Filter Design}).

\subsubsection{Pose-aware Magnetic Disturbance Detector}
\label{sec:Pose-aware Magnetic Field Selector}
In previous work, a common practice is to trigger magnetic field correction when the normalized magnitude of the magnetic field close to 1 \cite{zhang2012quaternion,sabatini2011estimating,madgwick2020extended}. However, even in a noisy magnetic environment, there are often instances where the magnetic field magnitude momentarily approaches 1. Our key observation is that, around these ``coincidental'' points, the magnetic field rarely maintains a consistent magnitude of 1 over space. 

Building on this observation, we propose a spatial consistency check for magnetic field correction (Fig. \ref{fig:detector}). For each IMU, we first identify its $k$ nearest neighboring IMUs based on the previously estimated pose and denote their indices as $N(i)$. We then verify whether the normalized magnitudes of their magnetic fields are all close to 1. If so, the magnetic field is deemed reliable and used for VO correction. We use $flag_i$ to indicate whether the $i$-th IMU should use magnetic field for VO correction, where True enables it and False disables it. Formally, this process can be described as follows:
\begin{equation}
  flag_i= 
\begin{cases}
\mathrm{True} &\text{if } \vert \Vert \boldsymbol m_\mathrm{S}^j \Vert -1 \vert < \epsilon_m, \forall j \in N(i), \\
\mathrm{False} & \text{otherwise},
\end{cases}
\end{equation}
here $i \in \{\mathrm{larm}, \dots, \mathrm{root}\}$, and $\epsilon_m$ stands for the threshold. In our implementation, we set $k=3$ and $\epsilon_m=0.15$. It should be noted that, according to our detection criteria, certain usable magnetic fields may be misclassified as adverse ones when in proximity to a disturbed magnetic field. We regard this phenomenon as a trade-off, which may sacrifice some accuracy. However, the cautious utilization of magnetic fields can enhance the stability of the system.

\subsubsection{Filter Design}
\label{sec:Filter Design}
With the $flag_i$ indicating whether to use magnetic field correction, each IMU individually fuses its raw measurements using an Error State Kalman Filter \cite{sola2017quaternion}. In the following derivation, we omit the IMU index $i$ for simplicity. The state space comprises orientation, bias of accelerometer and bias of gyroscope, denoted as $[\boldsymbol R_\mathrm{G}, \boldsymbol a_\mathrm{S}^\mathrm{bias}, \boldsymbol \omega_\mathrm{S}^\mathrm{bias}]$. In the prediction step, the angular velocity $\boldsymbol \omega_\mathrm{S}$ is used to update $\boldsymbol R_\mathrm{G}$ through integration. The discrete time prediction with time interval $\delta t$ can be expressed as: 

\begin{equation}
 \boldsymbol R_{\mathrm{G}, {t+1}} = \boldsymbol R_{\mathrm{G}, t} \mathrm {Exp} \left( (\boldsymbol \omega_{\mathrm{S}, t}-\boldsymbol\omega^\mathrm{bias}_{\mathrm{S}, t})\delta t \right),
\end{equation}
Here, $\mathrm {Exp}(\cdot)$ denotes the mapping from an axis-angle representation to a rotation matrix.
In the correction step, we apply two distinct correction methods, each triggered under specific conditions: VO of gravitational acceleration is activated when the acceleration magnitude is close to $9.8$, and VO of geomagnetic field is activated when the $flag$ is set to True. These constrains can be formulated as:
\begin{equation}
\boldsymbol g= \boldsymbol R_{\mathrm{G}, t}(\boldsymbol a_{\mathrm{S},t}-\boldsymbol a^\mathrm{bias}_{\mathrm{S},t}),
\end{equation}
\begin{equation}
\boldsymbol n= \boldsymbol R_{\mathrm{G}, t} \boldsymbol m_{\mathrm{S},t},
\label{eq:VO mag}
\end{equation}
Here $\boldsymbol g$ and $\boldsymbol n$ are the constant global gravitational acceleration and geomagnetic field, which are determined during initialization. $\epsilon_a$ is the threshold to select gravitational acceleration, which is set to $0.5$. Notably, before using Eq. \ref{eq:VO mag}, we project $\boldsymbol m_t$ to the plane orthogonal to $\boldsymbol g$, which ensures there's only ``yaw error'' \cite{shuster1981three}. Having defined the prediction and correction equations, the system state can be updated frame-by-frame following the ESKF workflow.

Now we have obtained the global orientation $\boldsymbol R_G$ for each IMU at each frame. Next, we can transform the sensor-local accelerations $\boldsymbol a_\mathrm{S}$ and angular velocities $\boldsymbol \omega_\mathrm{S}$ into the global frame using the global orientation. Finally, we obtaine $(\boldsymbol a, \boldsymbol \omega, \boldsymbol R)_\mathrm{G}^\mathrm{all}$.

\subsection{Neural Yaw Error Corrector}
\label{sec:Neural Yaw Error Corrector}
While the fusion process partially mitigates errors, in this stage we further eliminate the remaining errors using human motion prior. Specifically, when IMUs produce erroneous readings, these readings often correspond to implausible motions, such as joint rotations exceeding physiological limits or movements deviating from natural human motion patterns. This allows us to detect and correct such errors. However, correcting the yaw errors of 6 IMUs is an ambiguous task. For instance, if the readings from the root IMU and leaf IMUs are not consistent, it is difficult to determine whether the error occurs at the root or the leaf. Thus, in this stage, we assume the root IMU is correct and only estimate the relative errors of leaf IMUs. Formally, the inputs to this stage are global readings, $(\boldsymbol a, \boldsymbol \omega, \boldsymbol R)_\mathrm{G}^\mathrm{all}$, the outputs are corrected leaf IMUs readings $(\widetilde{ \boldsymbol a}, \widetilde {\boldsymbol \omega}, \widetilde {\boldsymbol R})_\mathrm{G}^{\mathrm{leaf}}$. 
In the following sections, we first introduce the network design for this task (Sec. \ref{sec:relative yaw error estimation}), then explain the synthesis of training data (Sec. \ref{sec:training data synthesis}) and finally describe how the network outputs are weighted to compute the correction values (Sec. \ref{sec:weighted correct strategy}).

\subsubsection{Relative Yaw Error Estimation}
\label{sec:relative yaw error estimation}
We use a neural network, named YawCorrector, to estimate the relative yaw error. The inputs to YawCorrector are the root-frame-related orientations and accelerations of 5 leaf IMUs as long as the gravity vector, denoted as $[{\boldsymbol{R}}_\mathrm{RL}, \boldsymbol{a}_\mathrm{RL}, \boldsymbol{g}_\mathrm R] \in \mathbb{R}^{5\times(9+3)+3}$. Notably, it is necessary to introduce the gravity vector in the input, which indicates along which axis the correction is made. Since IMUs only suffer yaw error, $\boldsymbol{g}_R$ is accurate most of the time. The outputs from YawCorrector are the relative error angles $\boldsymbol{\Delta} = [\Delta_{\mathrm{larm}}, ..., \Delta_{\mathrm{head}}] \in \mathbb{R}^{5}$. 

We use a standard LSTM \cite{hochreiter1997long} to regress the error angles, which provides two key advantages. First, by leveraging temporal context from multiple frames, it narrows the range of potential error angles. While a single frame allows for many plausible poses and a wide error range, sequential frames result in more precise predictions. Second, its recurrent structure ensures the continuity of the output angles, which directly translates to continuous corrected IMU readings, reducing jitter in the final motion capture results. The loss function is defined as:
\begin{equation}
 \mathcal{L} = \Vert \boldsymbol \Delta - \boldsymbol \Delta^{\mathrm{gt}} \Vert_2.
\end{equation}

\subsubsection{Training Data Synthesis}
\label{sec:training data synthesis}
To train our model, we require paired IMU readings with yaw error and corresponding error angles. However, datasets containing real IMU readings are scarce, and those with erroneous IMU readings are even rarer. To address this challenge, we develop a novel data synthesis technique that can simulate erroneous IMU readings affected by magnetic fields from MoCap sequences. Subsequently, the corresponding error angles can be easily obtained by comparing them with the ground truth rotations.

Our data synthesis method follows the pipeline of PNP \cite{PNP}, which first synthesizes raw measurements (accelerations, angular velocities, and magnetic field vectors) and then derives IMU readings through sensor fusion. The primary distinction lies in our approach to simulating magnetic fields: while PNP assumes a constant geomagnetic field, we introduce magnetic disturbances, thereby simulating the impact of magnetic interference on the IMU. Below, we detail our method for synthesizing magnetometer's measurements under magnetic interference.

We first generate a magnetic field by randomly placing magnets around a virtual room and then superimpose it with the Earth's magnetic field. Given a position $\boldsymbol x$, we can compute the \emph{global} magnetic field $\boldsymbol m_\mathrm{G}(\boldsymbol x)$ at that location using electromagnetic simulation \cite{ortner2020magpylib}. Then we put the MoCap character in the virtual environment. We use the standard mesh-based method \cite{DIP, loper2015smpl} to synthesize the 6DoF trajectory of each IMU from MoCap sequences \cite{mahmood2019amass}. We denote the obtained 6DoF trajectory as $\{\boldsymbol R, \boldsymbol p\}$. Finally we can compute the \emph{sensor-local} magnetometer's measurement through:
\begin{equation}
  \boldsymbol m_\mathrm S^{\mathrm{syn}} = \boldsymbol R^T m_\mathrm G(\boldsymbol p).
\end{equation}

The erroneous IMU orientation can be obtained through Error State Kalman Filter \cite{sola2017quaternion}:
\begin{equation}
  \boldsymbol R^\mathrm{err} = \mathrm{ESKF}(\boldsymbol a_\mathrm S^{\mathrm{syn}}, \boldsymbol \omega_\mathrm S^{\mathrm{syn}}, \boldsymbol m_\mathrm S^{\mathrm{syn}}),
\end{equation}
here $\boldsymbol a_\mathrm S^{\mathrm{syn}}$ and $\boldsymbol \omega_\mathrm S^{\mathrm{syn}}$ are synthesized following \cite{PNP}. The ground truth IMU orientation $\boldsymbol R$ has been already calculated when synthesize the 6DOF trajactory. By calculating the relative rotation between $\boldsymbol R^\mathrm{err}$ and $\boldsymbol R$ we obtain the absolute yaw error angles $\theta$ for each IMU. The relative error angles between the $i$-th leaf and the root can be calculated as:
\begin{equation}
  \Delta_i^\mathrm{gt} =\theta_i-\theta_\mathrm{root},
\end{equation}
which serve as supervision during the training process.

\subsubsection{Weighted Correct Strategy}
\label{sec:weighted correct strategy}
We then use the predicted error angle $\boldsymbol \Delta$ to correct leaf IMU readings $(\boldsymbol a, \boldsymbol \omega, \boldsymbol R)_\mathrm{G}^\mathrm{leaf}$. This process corresponds to the Motion Prior Correction shown in the Fig. \ref{fig:pipeline}. However, the output of YawCorrector is not directly used to correct IMU readings but is instead weighted. Intuitively, when the ambient magnetic field is reliable, we tend to avoid using the YawCorrector. Conversely, when the magnetic field is unreliable, we prefer to employ the YawCorrector. Specifically, we define a weight $w \in [0, 1]$. It is initialized by 0 and it is updated according to the following equation in each frame:
\begin{equation}
  w=
\begin{cases}
w - 0.05 & \text{if } \cap flag_i,\\
w + 0.05 & \text{otherwise},
\end{cases}
\end{equation}
$w$ is always clamped between $[0, 1]$. This update formula means: if no IMU detects magnetic disturbance in a given frame, we reduce the weight; once any IMU detects magnetic disturbance, we increase the weight.

We use the weighted output to correct the leaf IMUs by:
\begin{equation}
  \widetilde{\boldsymbol R_\mathrm G^i} = \boldsymbol{\mathrm R}_\mathrm{g}(-w\Delta_i)  \boldsymbol R_\mathrm G^i,
\end{equation}
\begin{equation}
  \widetilde{\boldsymbol a_\mathrm G^i} = \boldsymbol{\mathrm R}_\mathrm{g}(-w\Delta_i)  \boldsymbol a_\mathrm G^i,
\end{equation}
\begin{equation}
  \widetilde{\boldsymbol {\omega}_\mathrm G^i} = \boldsymbol{\mathrm R}_\mathrm{g}(-w\Delta_i)  \boldsymbol \omega_\mathrm G^i,
\end{equation}
where $\boldsymbol{\mathrm R}_\mathrm{g}(\theta)$ donates a rotation of $\theta$ radians around the gravity axis, and 
$i \in \{\mathrm{larm}, \dots, \mathrm{head}\}$ represents the $i$-th leaf joint. Finally, the corrected leaf IMU readings $(\widetilde{ \boldsymbol a}, \widetilde {\boldsymbol \omega}, \widetilde {\boldsymbol R})_\mathrm{G}^{\mathrm{leaf}}$, together with the root IMU readings $(\boldsymbol a, \boldsymbol \omega, \boldsymbol m)_\mathrm{G}^{\mathrm{root}}$, are fed to some spare inertial MoCap system.

\section{Experiments}
In this section, we first present the implementation details (Sec. \ref{sec:impl detail}), and then conduct comparisons with previous methods (Sec. \ref{sec: comp}). We do further ablation studies to evaluate our key designs (Sec. \ref{sec:abl}). At last, we discuss our limitations (Sec. \ref{sec:limit}).

\subsection{Implementation Details}
\label{sec:impl detail}
\paragraph{Network Structure.}
Our YawCorrector adopts an RNN architecture, comprising a linear input layer, two Long Short-term Memory (LSTM) \cite{hochreiter1997long} layers, and a linear output layer. The linear layers employ ReLU as their activation function, and the hidden dimension is set to 256. During training, the dropout rate is set to 40\%, batch size is set to 256, and Adam optimizer \cite{KingmaB14} is used.

\paragraph{Hardware and Performance}
We conduct performance tests on Intel(R) Core(TM) i9-13900KF CPU and an NVIDIA RTX4090 GPU. Our sensor fusion can run at 1000fps and YawCorrector can run at 400fps, indicating that our method is lightweight enough.
When the most time consuming sparse inertial MoCap system PNP \cite{PNP} is integrated with our method, it can still run at 60fps. In live demo, we use PN Lab sensors from Noitom \cite{noitom} which transmits raw measurements at 100fps.

\paragraph{Datasets.} For training, we only use synthesized data from AMASS \cite{mahmood2019amass}. For testing, since previous datasets are collected in environments with minimal magnetic interference, we collect a new dataset under severe magnetic interference, named MagIMU. We use MagIMU to evaluate our method's performance under magnetic disturbances and use TotalCapture \cite{trumble2017total} to evaluate its performance in environments with clean magnetic fields. Some basic information of MagIMU dataset is as follows. The dataset contains approximately 70 minutes of motion data captured from 5 subjects, each equipped with 6 IMUs, and includes raw IMU measurements at 100fps, SMPL poses, and global translations. For more detailed information, please refer to the supplementary material.

\paragraph{Evaluation Principles.}
Although our method is responsible for IMU orientation estimation, we do not directly evaluate the accuracy of IMU orientations in our experiments. This is due to two reasons: (1) obtaining ground truth for IMU orientations is challenging, and (2) in motion capture scenarios, it is essential to consider not only the accuracy of individual IMUs but also their coordination. Therefore, we alternatively evaluate an IMU orientation estimation method by feeding their output IMU readings to inertial posers and assessing the accuracy of the final MoCap results. Two state-of-the-art sparse inertial posers are used for evaluation process: PNP \cite{PNP} and DynaIP \cite{DynaIP}. PNP combines data-driven and physics-driven techniques, jointly estimating pose and translation, while DynaIP is a purely data-driven approach that only estimates pose.

\paragraph{Metrics.} To evaluate the accuracy of local pose, we use the following metrics: (1) SIP Error ($^\circ$): the mean global rotation error of the hips and shoulders; (2) Angular Error ($^\circ$): the mean global rotation error of all joints; (3) Positional Error (cm): the mean position error of all joints; (4) Mesh Error (cm): the mean position error of the SMPL mesh vertices. Before measuring these metrics, we first align the root joint of the estimated pose with the ground truth. To evaluate the accuracy of translation, we calculate the cumulative translation error. Specifically, we first divide the estimated and ground truth motion trajectories into segments based on the subject's actual travel distance. For each segment, we align the position and heading of the first frame and then compute the positional error of the last frame.

\paragraph{Baseline.} We propose an IMU orientation estimation method specifically designed for sparse inertial motion capture scenarios. To demonstrate its capability, we establish a traditional fusion-based method as the baseline, which detects magnetic field disturbances \emph{individually} for each IMU. In other words, it can be regarded as a special case of our method proposed in Sec. \ref{sec:Pose-aware Multiple IMUs Fusion} when $k=1$. Notably, this baseline method performs comparably to the commercial IMU \cite{noitom} in terms of motion capture accuracy. The comparison between them is shown in supplementary material.

\begin{figure}
    \centering
    \includegraphics[width=1\linewidth]{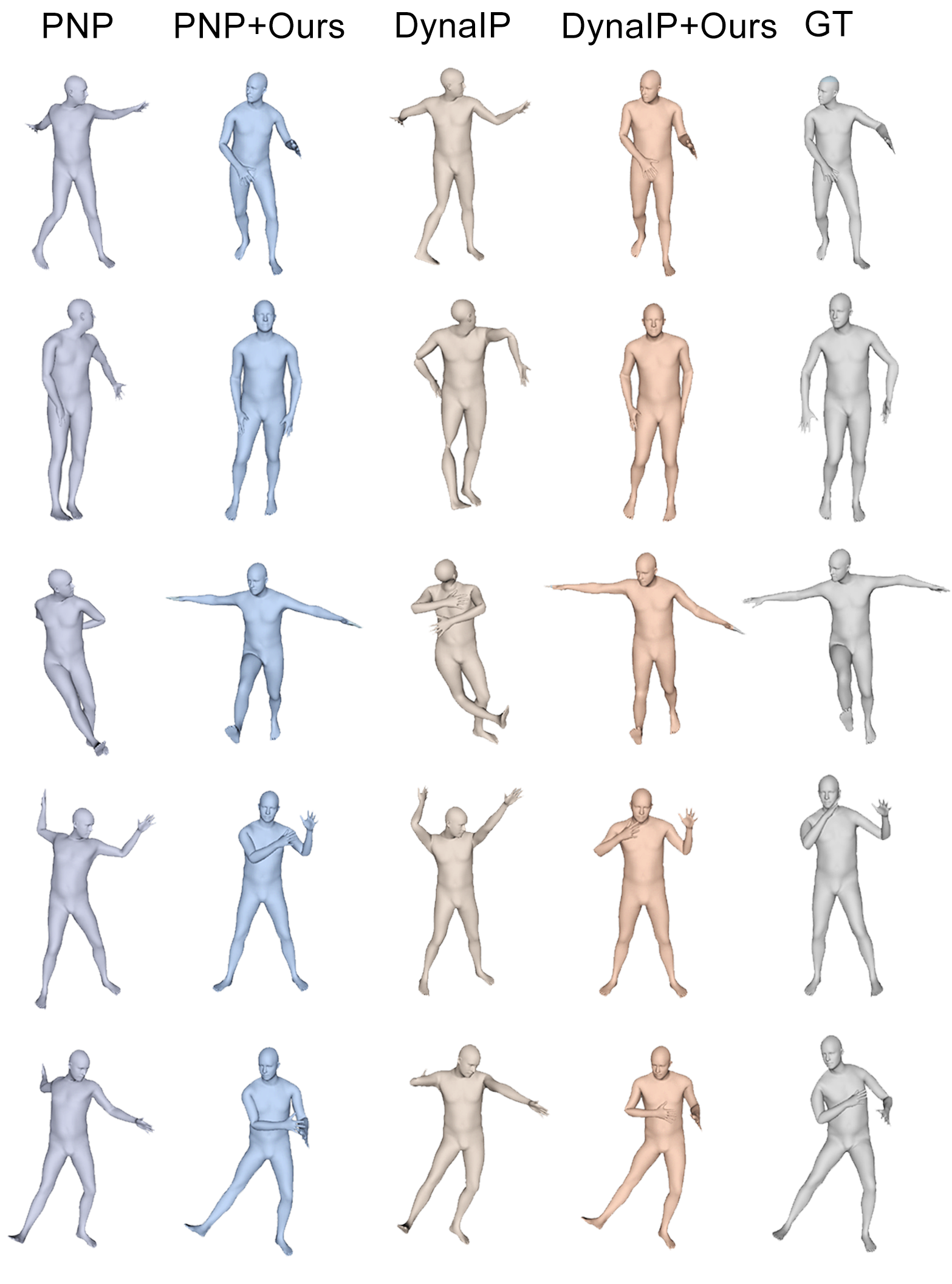}
    \caption{Qualitative comparison with baseline methods. Examples are picked from the newly collected MagIMU dataset. }
    \label{fig:qualitative}
\end{figure}

\subsection{Comparisons}
\label{sec: comp}
We compare our method with the baseline on MagIMU from two perspectives: local pose and global translation.

As for local pose estimation, the quantitative results are shown in Tab. \ref{tab:Comparisons}. Comparing with baseline, our method reduces errors by about 10\% in all metrics, which comes from our method's ability to mitigate magnetic disturbances. Among these metrics, joint rotation error shows the most significant reduction, reaching 25\%, while the improvements in other metrics are less pronounced. This can be attributed to two main reasons. First, yaw error often does not lead to significant deviations in joint or vertex positions. For instance, when a person stands upright, yaw error may cause the arms and legs to severely twist or contort, yet the joint and vertex remain almost unchanged. Second, yaw errors primarily occur at the leaf nodes, where they tend to cancel each other out, resulting in minimal impact on the torso. As a result, the SIP error, which measures torso pose accuracy, shows little change. In short, we argue that angular error better reflects the resilience to magnetic disturbances. 
Additionally, our method demonstrates effectiveness across the two inertial posers, indicating its compatibility with different sparse inertial MoCap systems. The qualitative results are shown in Fig. \ref{fig:qualitative}. When the baseline method is used as the system input, these sparse inertial MoCap systems are significantly affected by the so-called ``yaw error'' and output implausible results. Meanwhile, when integrated with our method, the results become much more accurate.

\begin{figure}
    \centering
    \includegraphics[width=\linewidth]{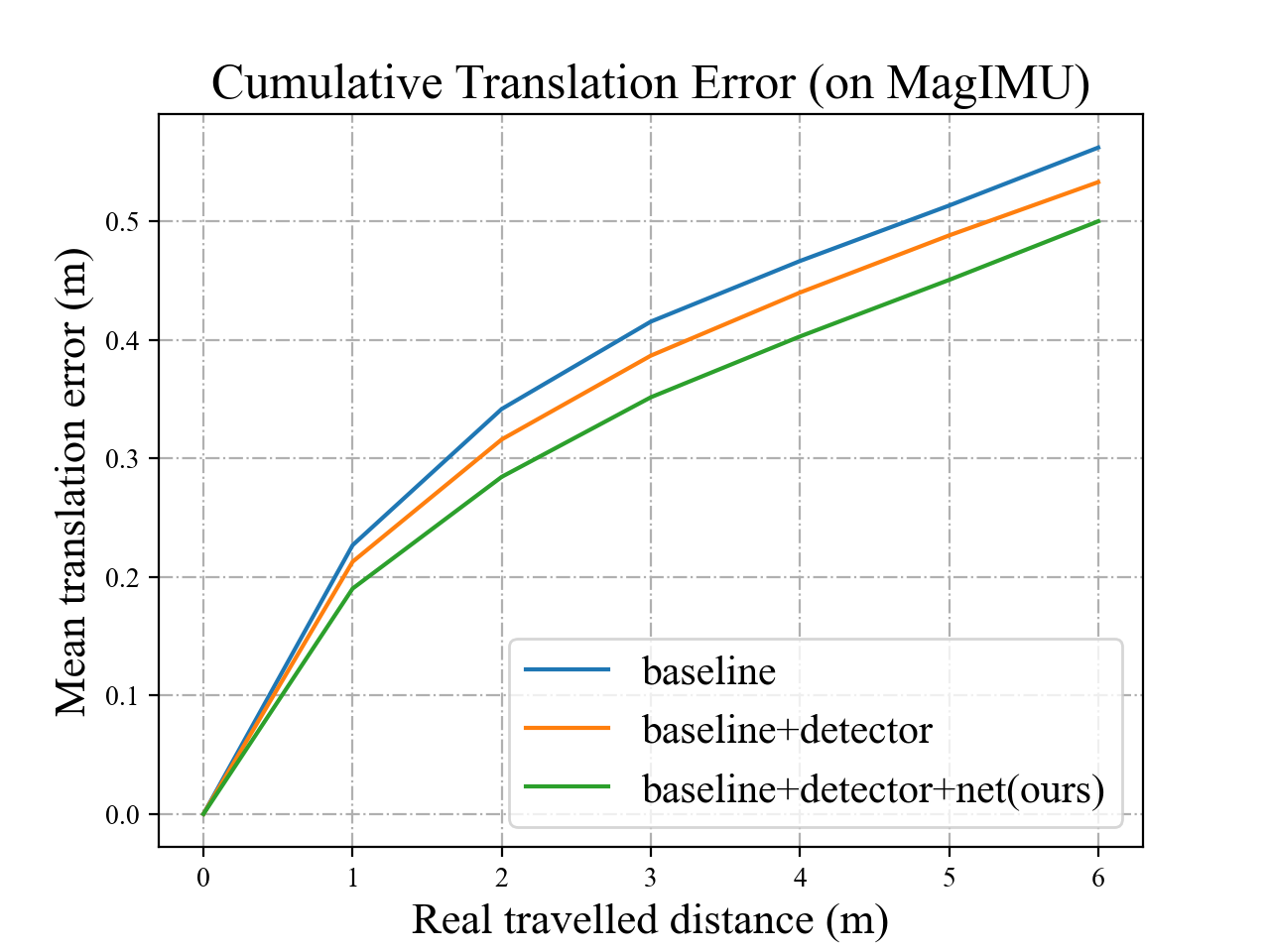}
    \caption{Translation comparisons of different IMU estimation methods integrated into the PNP pipeline. Lower curve indicates smaller error.}
    \label{fig:ablation_tran}
\end{figure}

We also compare the translation error. Since DynaIP \cite{DynaIP} doesn't estimate global translation, we only conduct comparison on PNP \cite{PNP}. As shown in Fig. \ref{fig:ablation_tran}, our method has lower global position error, which comes from the more accurate IMUs readings.

\begin{table}
  \centering
  \setlength{\tabcolsep}{3pt}
  \begin{tabular}{@{}lcccc@{}}
    \toprule
    Method & SIP Err & Ang Err & Pos Err &  Mesh Err\\
    \midrule
    \multicolumn{5}{c}{PNP} \\ 
    \midrule
    baseline &  26.63 & 24.65 & 8.99 & 10.86\\
    +detector & 25.67 & 23.20 & 8.56 & 10.26 \\
    +detector+net (ours) & \pmb{24.19} & \pmb{20.23} & \pmb{8.17} & \pmb{9.74}\\
    \midrule
    \multicolumn{5}{c}{DynaIP} \\ 
    \midrule
    baseline &  31.55 & 28.79 & 9.10 & 11.17\\
    +detector &  30.79 & 27.26 & 8.64 & 10.67\\
    +detector+net (ours) & \pmb{28.68} & \pmb{22.12} & \pmb{8.05} & \pmb{9.84}\\ 
    \bottomrule
  \end{tabular}
  \caption{Quantitative comparisons of different IMU estimation methods on local pose. Conducted on MagIMU.}
  \label{tab:Comparisons}
\end{table}

\subsection{Ablation Studys}
\label{sec:abl}
\paragraph{Key Components.} Our method is composed of two key components: a pose-aware magnetic disturbance detector (referred to as the detector) and a neural yaw error corrector (referred to as the net). To further illustrate the function of each components, we compare three methods: baseline, baseline+detector, baseline+detector+net. We don't compare baseline+net with them since our YawCorrector net needs the information from the detector to determine its correcting weight. From Tab. \ref{tab:Comparisons} and Fig. \ref{fig:ablation_tran}, we can clearly see that both the detector and the net lead to improvements. It is worth noting that although the YawCorrector net only corrects leaf nodes, it still contributes to global translation. This is because, in PNP, global velocity estimation relies on local pose, and more accurate local poses lead to more accurate global velocities. Additionally, local poses also influence global translation through ground contact constraints.

\paragraph{Erroneous IMU synthesis.} 
When preparing training data, we synthesize IMU readings with yaw error by add noise to magnetic field and simulate its impact on IMU readings through sensor fusion. To demonstrate the benefit of our approach, we compare it with a naive approach. In the naive approach, we first synthesize IMU readings without magnetic disturbance, and then add walking noise along yaw direction to get erroneous IMU readings. Using these data, we train another YawCorrector, and compare its performance with our version. As reported in Tab. \ref{tab:Ablation_SynData}, when our data synthesis method was replaced by the naive one, the trained model exhibited a slight decline in performance. We suggest that the advantage of our method stems from its more realistic simulation of IMU behavior under real-world magnetic disturbances.

\paragraph{Performance on clean data.} We also evaluate our method on TotalCapture \cite{trumble2017total} dataset to ensure that it does not introduce adverse effects in environments with minimal magnetic interference. Since it only contains data in 60 fps, we can't run our multi-IMU fusion algorithm which involve high frequency integration. Thus, we replace it with official IMU readings, but keep the pose-aware magnetic field detector for determining the weight for YawCorrector. we conduct the comparison between origin PNP and PNP enhanced by our method. As the result, our method achieves performance comparable to previous methods, indicating that our method can serve as an always-on component across different environments. Since the results show no significant differences, we present them in the supplementary material.

\begin{table}
  \centering
  \begin{tabular}{@{}lcccc@{}}
    \toprule
    Method & SIP Err & Ang Err & Pos Err &  Mesh Err\\
    \midrule
    \multicolumn{5}{c}{PNP} \\ 
    \midrule
    +w/o syn mf&  24.41 & 20.53 & 8.26 & 9.86\\
    +ours & \pmb{24.19} & \pmb{20.23} & \pmb{8.17} & \pmb{9.74} \\
    \midrule
    \multicolumn{5}{c}{DynaIP} \\ 
    \midrule
    +w/o syn mf &  28.78 & 22.51 & 8.12 & 9.91 \\
    +ours & \pmb{28.68} & \pmb{22.12} & \pmb{8.05} & \pmb{9.84} \\
    \bottomrule
  \end{tabular}
  \caption{Evaluation on our proposed IMU synthesize approach that first synthesis magnetic field (syn mf). Conducted on MagIMU.}
  \label{tab:Ablation_SynData}
\end{table}

\subsection{Limitation}
\label{sec:limit}
Our proposed pose-aware magnetic disturbance detector utilizes only the positional information of IMUs and does not incorporate their relative rotations. Combining the relative rotations between IMUs and the local magnetic field directions at each IMU could potentially lead to better results. Additionally, our method does not account for the issue of magnetometer magnetization, which may lead to significant errors in magnetometer measurements. In such cases, our approach could potentially fail.

\section{Conclusion}
This work proposes a novel method to address magnetic interference in sparse inertial MoCap systems. To the best of our knowledge, it is the first work dealing with the resilience of sparse inertial MoCap systems to magnetic disturbances. On one hand, we propose a pose-aware magnetic disturbance detector that prevents IMUs from being misled by non-geomagnetic fields by aggregating information from neighboring IMUs. On the other hand, we introduce a human motion prior-based IMU correction network that detects and corrects errors in IMU readings, providing a post-hoc remedy. Experiments demonstrate that our method enhances the performance of various sparse Inertial Mocap systems in magnetically disturbed environments.
{
    \small
    \bibliographystyle{ieeenat_fullname}
    \bibliography{main}

\begin{thebibliography}{35}
\providecommand{\natexlab}[1]{#1}
\providecommand{\url}[1]{\texttt{#1}}
\expandafter\ifx\csname urlstyle\endcsname\relax
  \providecommand{\doi}[1]{doi: #1}\else
  \providecommand{\doi}{doi: \begingroup \urlstyle{rm}\Url}\fi

\bibitem[Bachmann et~al.(2004)Bachmann, Yun, and Peterson]{bachmann2004investigation}
Eric~R Bachmann, Xiaoping Yun, and Christopher~W Peterson.
\newblock An investigation of the effects of magnetic variations on inertial/magnetic orientation sensors.
\newblock In \emph{IEEE International Conference on Robotics and Automation, 2004. Proceedings. ICRA'04. 2004}, pages 1115--1122. IEEE, 2004.

\bibitem[Bhardwaj et~al.(2018)Bhardwaj, Kumar, and Kumar]{bhardwaj2018errors}
Renu Bhardwaj, Neelesh Kumar, and Vipan Kumar.
\newblock Errors in micro-electro-mechanical systems inertial measurement and a review on present practices of error modelling.
\newblock \emph{Transactions of the Institute of Measurement and Control}, 40\penalty0 (9):\penalty0 2843--2854, 2018.

\bibitem[Fan et~al.(2017{\natexlab{a}})Fan, Li, and Liu]{fan2017magnetic}
Bingfei Fan, Qingguo Li, and Tao Liu.
\newblock How magnetic disturbance influences the attitude and heading in magnetic and inertial sensor-based orientation estimation.
\newblock \emph{Sensors}, 18\penalty0 (1):\penalty0 76, 2017{\natexlab{a}}.

\bibitem[Fan et~al.(2017{\natexlab{b}})Fan, Li, Wang, and Liu]{fan2017adaptive}
Bingfei Fan, Qingguo Li, Chao Wang, and Tao Liu.
\newblock An adaptive orientation estimation method for magnetic and inertial sensors in the presence of magnetic disturbances.
\newblock \emph{Sensors}, 17\penalty0 (5):\penalty0 1161, 2017{\natexlab{b}}.

\bibitem[Hochreiter and Schmidhuber(1997)]{hochreiter1997long}
Sepp Hochreiter and J{\"u}rgen Schmidhuber.
\newblock Long short-term memory.
\newblock \emph{Neural computation}, 9\penalty0 (8):\penalty0 1735--1780, 1997.

\bibitem[Huang et~al.(2018)Huang, Kaufmann, Aksan, Black, Hilliges, and Pons-Moll]{DIP}
Yinghao Huang, Manuel Kaufmann, Emre Aksan, Michael~J Black, Otmar Hilliges, and Gerard Pons-Moll.
\newblock Deep inertial poser: Learning to reconstruct human pose from sparse inertial measurements in real time.
\newblock \emph{ACM Transactions on Graphics (TOG)}, 37\penalty0 (6):\penalty0 1--15, 2018.

\bibitem[Jiang et~al.(2022)Jiang, Ye, Gopinath, Won, Winkler, and Liu]{TIP}
Yifeng Jiang, Yuting Ye, Deepak Gopinath, Jungdam Won, Alexander~W Winkler, and C~Karen Liu.
\newblock Transformer inertial poser: Real-time human motion reconstruction from sparse imus with simultaneous terrain generation.
\newblock In \emph{SIGGRAPH Asia 2022 Conference Papers}, pages 1--9, 2022.

\bibitem[Kingma and Ba(2015)]{KingmaB14}
Diederik~P. Kingma and Jimmy Ba.
\newblock Adam: {A} method for stochastic optimization.
\newblock In \emph{3rd International Conference on Learning Representations, {ICLR} 2015, San Diego, CA, USA, May 7-9, 2015, Conference Track Proceedings}, 2015.

\bibitem[Laidig and Seel(2023)]{laidig2023vqf}
Daniel Laidig and Thomas Seel.
\newblock Vqf: Highly accurate imu orientation estimation with bias estimation and magnetic disturbance rejection.
\newblock \emph{Information Fusion}, 91:\penalty0 187--204, 2023.

\bibitem[Lee and Park(2009)]{lee2009minimum}
Jung~Keun Lee and Edward~J Park.
\newblock Minimum-order kalman filter with vector selector for accurate estimation of human body orientation.
\newblock \emph{IEEE Transactions on Robotics}, 25\penalty0 (5):\penalty0 1196--1201, 2009.

\bibitem[Loper et~al.(2015)Loper, Mahmood, Romero, Pons-Moll, and Black]{loper2015smpl}
Matthew Loper, Naureen Mahmood, Javier Romero, Gerard Pons-Moll, and Michael~J Black.
\newblock Smpl: A skinned multi-person linear model.
\newblock \emph{ACM Transactions on Graphics}, 34\penalty0 (6), 2015.

\bibitem[Madgwick et~al.(2020)Madgwick, Wilson, Turk, Burridge, Kapatos, and Vaidyanathan]{madgwick2020extended}
Sebastian~OH Madgwick, Samuel Wilson, Ruth Turk, Jane Burridge, Christos Kapatos, and Ravi Vaidyanathan.
\newblock An extended complementary filter for full-body marg orientation estimation.
\newblock \emph{IEEE/ASME Transactions on mechatronics}, 25\penalty0 (4):\penalty0 2054--2064, 2020.

\bibitem[Mahmood et~al.(2019)Mahmood, Ghorbani, Troje, Pons-Moll, and Black]{mahmood2019amass}
Naureen Mahmood, Nima Ghorbani, Nikolaus~F Troje, Gerard Pons-Moll, and Michael~J Black.
\newblock Amass: Archive of motion capture as surface shapes.
\newblock In \emph{Proceedings of the IEEE/CVF international conference on computer vision}, pages 5442--5451, 2019.

\bibitem[Mollyn et~al.(2023)Mollyn, Arakawa, Goel, Harrison, and Ahuja]{IMUPoser}
Vimal Mollyn, Riku Arakawa, Mayank Goel, Chris Harrison, and Karan Ahuja.
\newblock Imuposer: Full-body pose estimation using imus in phones, watches, and earbuds.
\newblock In \emph{Proceedings of the 2023 CHI Conference on Human Factors in Computing Systems}, pages 1--12, 2023.

\bibitem[Noitom(2025)]{noitom}
Noitom.
\newblock Noitom motion capture systems, 2025.

\bibitem[Ortner and Bandeira(2020)]{ortner2020magpylib}
Michael Ortner and Lucas Gabriel~Coliado Bandeira.
\newblock Magpylib: A free python package for magnetic field computation.
\newblock \emph{SoftwareX}, 11:\penalty0 100466, 2020.

\bibitem[Roetenberg et~al.(2005)Roetenberg, Luinge, Baten, and Veltink]{roetenberg2005compensation}
Daniel Roetenberg, Henk~J Luinge, Chris~TM Baten, and Peter~H Veltink.
\newblock Compensation of magnetic disturbances improves inertial and magnetic sensing of human body segment orientation.
\newblock \emph{IEEE Transactions on neural systems and rehabilitation engineering}, 13\penalty0 (3):\penalty0 395--405, 2005.

\bibitem[Sabatini(2011)]{sabatini2011estimating}
Angelo~Maria Sabatini.
\newblock Estimating three-dimensional orientation of human body parts by inertial/magnetic sensing.
\newblock \emph{Sensors}, 11\penalty0 (2):\penalty0 1489--1525, 2011.

\bibitem[Sabatini(2012)]{sabatini2012variable}
Angelo~Maria Sabatini.
\newblock Variable-state-dimension kalman-based filter for orientation determination using inertial and magnetic sensors.
\newblock \emph{Sensors}, 12\penalty0 (7):\penalty0 8491--8506, 2012.

\bibitem[Shuster and Oh(1981)]{shuster1981three}
Malcolm~David Shuster and S~D\_ Oh.
\newblock Three-axis attitude determination from vector observations.
\newblock \emph{Journal of guidance and Control}, 4\penalty0 (1):\penalty0 70--77, 1981.

\bibitem[Sola(2017)]{sola2017quaternion}
Joan Sola.
\newblock Quaternion kinematics for the error-state kalman filter.
\newblock \emph{arXiv preprint arXiv:1711.02508}, 2017.

\bibitem[Suh et~al.(2012)Suh, Ro, and Kang]{suh2012quaternion}
Young~Soo Suh, Young~Shick Ro, and Hee~Jun Kang.
\newblock Quaternion-based indirect kalman filter discarding pitch and roll information contained in magnetic sensors.
\newblock \emph{IEEE Transactions on Instrumentation and measurement}, 61\penalty0 (6):\penalty0 1786--1792, 2012.

\bibitem[Trumble et~al.(2017)Trumble, Gilbert, Malleson, Hilton, and Collomosse]{trumble2017total}
Matthew Trumble, Andrew Gilbert, Charles Malleson, Adrian Hilton, and John Collomosse.
\newblock Total capture: 3d human pose estimation fusing video and inertial sensors.
\newblock In \emph{Proceedings of 28th British Machine Vision Conference}, pages 1--13, 2017.

\bibitem[Van~Wouwe et~al.(2024)Van~Wouwe, Lee, Falisse, Delp, and Liu]{DiffPoser}
Tom Van~Wouwe, Seunghwan Lee, Antoine Falisse, Scott Delp, and C~Karen Liu.
\newblock Diffusionposer: Real-time human motion reconstruction from arbitrary sparse sensors using autoregressive diffusion.
\newblock In \emph{Proceedings of the IEEE/CVF Conference on Computer Vision and Pattern Recognition}, pages 2513--2523, 2024.

\bibitem[Von~Marcard et~al.(2017)Von~Marcard, Rosenhahn, Black, and Pons-Moll]{SIP}
Timo Von~Marcard, Bodo Rosenhahn, Michael~J Black, and Gerard Pons-Moll.
\newblock Sparse inertial poser: Automatic 3d human pose estimation from sparse imus.
\newblock In \emph{Computer graphics forum}, pages 349--360. Wiley Online Library, 2017.

\bibitem[Wu et~al.(2025)Wu, Yin, Guo, Qin, et~al.]{ASIP}
Yinghao Wu, Lu Yin, Shihui Guo, Yipeng Qin, et~al.
\newblock Accurate and steady inertial pose estimation through sequence structure learning and modulation.
\newblock \emph{Advances in Neural Information Processing Systems}, 37:\penalty0 42468--42493, 2025.

\bibitem[Xsens(2025)]{xsens}
Xsens.
\newblock Xsens motion capture solutions, 2025.

\bibitem[Xu et~al.(2024)Xu, Gao, Hoffmann, and Ahuja]{MobilePoser}
Vasco Xu, Chenfeng Gao, Henry Hoffmann, and Karan Ahuja.
\newblock Mobileposer: Real-time full-body pose estimation and 3d human translation from imus in mobile consumer devices.
\newblock In \emph{Proceedings of the 37th Annual ACM Symposium on User Interface Software and Technology}, pages 1--11, 2024.

\bibitem[Yadav and Bleakley(2014)]{yadav2014accurate}
Nagesh Yadav and Chris Bleakley.
\newblock Accurate orientation estimation using ahrs under conditions of magnetic distortion.
\newblock \emph{Sensors}, 14\penalty0 (11):\penalty0 20008--20024, 2014.

\bibitem[Yi et~al.(2021)Yi, Zhou, and Xu]{TransPose}
Xinyu Yi, Yuxiao Zhou, and Feng Xu.
\newblock Transpose: Real-time 3d human translation and pose estimation with six inertial sensors.
\newblock \emph{ACM Transactions On Graphics (TOG)}, 40\penalty0 (4):\penalty0 1--13, 2021.

\bibitem[Yi et~al.(2022)Yi, Zhou, Habermann, Shimada, Golyanik, Theobalt, and Xu]{PIP}
Xinyu Yi, Yuxiao Zhou, Marc Habermann, Soshi Shimada, Vladislav Golyanik, Christian Theobalt, and Feng Xu.
\newblock Physical inertial poser (pip): Physics-aware real-time human motion tracking from sparse inertial sensors.
\newblock In \emph{Proceedings of the IEEE/CVF conference on computer vision and pattern recognition}, pages 13167--13178, 2022.

\bibitem[Yi et~al.(2024)Yi, Zhou, and Xu]{PNP}
Xinyu Yi, Yuxiao Zhou, and Feng Xu.
\newblock Physical non-inertial poser (pnp): modeling non-inertial effects in sparse-inertial human motion capture.
\newblock In \emph{ACM SIGGRAPH 2024 Conference Papers}, pages 1--11, 2024.

\bibitem[Zhang et~al.(2016)Zhang, Yu, Liu, Yuan, and Liu]{zhang2016dual}
Shengzhi Zhang, Shuai Yu, Chaojun Liu, Xuebing Yuan, and Sheng Liu.
\newblock A dual-linear kalman filter for real-time orientation determination system using low-cost mems sensors.
\newblock \emph{Sensors}, 16\penalty0 (2):\penalty0 264, 2016.

\bibitem[Zhang et~al.(2024)Zhang, Xia, Chu, Yang, Wu, and Pei]{DynaIP}
Yu Zhang, Songpengcheng Xia, Lei Chu, Jiarui Yang, Qi Wu, and Ling Pei.
\newblock Dynamic inertial poser (dynaip): Part-based motion dynamics learning for enhanced human pose estimation with sparse inertial sensors.
\newblock In \emph{Proceedings of the IEEE/CVF Conference on Computer Vision and Pattern Recognition}, pages 1889--1899, 2024.

\bibitem[Zhang et~al.(2012)Zhang, Meng, and Wu]{zhang2012quaternion}
Zhi-Qiang Zhang, Xiao-Li Meng, and Jian-Kang Wu.
\newblock Quaternion-based kalman filter with vector selection for accurate orientation tracking.
\newblock \emph{IEEE Transactions on Instrumentation and Measurement}, 61\penalty0 (10):\penalty0 2817--2824, 2012.

\end{thebibliography}
}

\end{document}